\documentclass[conference]{IEEEtran}
\IEEEoverridecommandlockouts
\usepackage{cite}
\usepackage{amsmath,amssymb,amsfonts}
\usepackage{algorithmic}
\usepackage{graphicx}
\usepackage{textcomp}
\usepackage{xcolor}
\usepackage{hyperref}
\usepackage{blindtext}

\hypersetup{
 colorlinks=true,      
 urlcolor=blue,
}

\def\BibTeX{{\rm B\kern-.05em{\sc i\kern-.025em b}\kern-.08em
    T\kern-.1667em\lower.7ex\hbox{E}\kern-.125emX}}
\begin{document}

\title{ANNA: A Deep Learning Based Dataset in Heterogeneous Traffic for Autonomous Vehicles\\
%  \thanks{Identify applicable funding agency here. If none, delete this.}
}

\author{
\IEEEauthorblockN{Mahedi Kamal\textsuperscript{*1}\thanks{\noindent\textsuperscript{*}These authors contributed equally to this work}\, 
Tasnim Fariha\textsuperscript{*1},
Afrina Kabir Zinia\textsuperscript{1},\\ Md. Abu Syed \textsuperscript{1}, Fahim Hasan Khan\textsuperscript{2}, Md. Mahbubur Rahman\textsuperscript{1} }

\IEEEauthorblockA{\textsuperscript{1}Department of Computer Science and Engineering, Military Institute of Science and Technology,\\ Mirpur Cantonment, Dhaka-1216, Bangladesh\\}

\IEEEauthorblockA{\textsuperscript{2}University of California, Santa Cruz, CA 95064, USA\\}

\IEEEauthorblockA{Email: mahedikamal44@gmail.com, nirjonafariha@gmail.com, afrinakabir104@gmail.com,\\ nihonsyed@gmail.com, fkhan4@ucsc.edu, mahbub@cse.mist.ac.bd}
 
}

% \and
% \and
% \and
% \and
% \and

\maketitle

\begin{abstract}
Recent breakthroughs in artificial intelligence offer tremendous promise for the development of self-driving applications. Deep Neural Networks, in particular, are being utilized to support the operation of semi-autonomous cars through object identification and semantic segmentation. To assess the inadequacy of the current dataset in the context of autonomous and semi-autonomous cars, we created a new dataset named ANNA. This study discusses a custom-built dataset that includes some unidentified vehicles in the perspective of Bangladesh, which are not included in the existing dataset. A dataset validity check was performed by evaluating models using the Intersection Over Union (IOU) metric. The results demonstrated that the model trained on our custom dataset was more precise and efficient than the models trained on the KITTI or COCO dataset concerning Bangladeshi traffic. The research presented in this paper also emphasizes the importance of developing accurate and efficient object detection algorithms for the advancement of autonomous vehicles.

%\footnotetext{Mahedi Kamal and Tasnim Fariha contributed equally}
\end{abstract}


\section{Introduction}
The advancement of general-purpose computers in the late 1990s made it possible to process large amounts of data quickly. A common approach at that time involved extracting feature vectors from images and applying machine learning techniques for image recognition. However, the emergence of deep learning revolutionized computer vision and greatly enhanced its capabilities. Computer vision, an area of Artificial Intelligence (AI), enables computers to perceive and understand digital images in a manner similar to human vision. This technology finds applications in various industries, such as facial identification, augmented reality, driverless cars, healthcare, and more. For autonomous vehicles (AVs) to operate effectively, they require accurate perception of their surroundings, which can be achieved through computer vision technology.

As Bangladesh is a densely populated country with congested roads, the viability of autonomous cars in such an environment is uncertain. Therefore, we aim to obtain a fresh dataset from the roadways of Bangladesh and conduct a comparative analysis with existing datasets. Our objective is to develop a real-time object detection Deep Learning (DL) model that can run directly on mobile phones without requiring server-side processing or an internet connection ~\cite{khan2021}. This model is trained on the new dataset, which includes objects and vehicles specific to Bangladesh and not present in resources like Microsoft COCO or other AV datasets. By comparing the MS COCO dataset with our new dataset, we aim to assess the performance of the AV model on the new dataset and determine the potential benefits of autonomous vehicles on Bangladeshi roads.

Our primary focus is on data collection with the detection of multiple objects on roadways for autonomous vehicles utilizing real-time object detection modules.  The Waymo Open Dataset Challenges are accepted which contain "Real-time 2D Detection" ~\cite{p43, waymo2021} and "3D Camera-Only Detection" ~\cite{waymo2022}. All prior Waymo Open Dataset challenges used data from multiple cameras, LiDAR, Radar, and other sensors, until Tesla popularised the notion of camera-only detection for AV ~\cite{p45}. Because Tesla and Waymo are moving in that route, our focus is solely on vision only using smartphones and all other costly hardware components are abandoned.

Multiple object detection on a mobile camera for AV necessitates detecting and identifying multiple things in the camera's field of view using computer vision techniques. Other cars, pedestrians, traffic signs, and other entities that may be significant to the vehicle's navigation and object avoidance can be included. The autonomous car then uses this information to assess its surroundings, plot a course, and make judgments while driving. This data is used by the control system of driverless vehicles to make safe and efficient driving decisions. YOLOv5 (You Only Look Once version 5) is a cutting-edge object detection method used in computer vision. Using this model is intended for speed and accuracy optimization in real-world object identification applications, and it employs a single CNN to recognize items inside an image and calculate the bounding box coordinates and class probabilities for each identified object~\cite{p41, p18}. On standard datasets, YOLOv5 has proven great accuracy and high-speed detection in contrast to other models, and its compact model size makes it suitable for usage on devices with limited computational resources~\cite{p46, p47}. In this research, we concentrated on collecting image data from public roads using only a single smartphone camera. We assembled and annotated a dataset utilizing images of real-world photographs extracted from videos that were shot from Bangladeshi urban streets. We have tested the dataset to evaluate the performance of the CNN-based YOLOv5 model for identifying unique vehicles which are available in Bangladesh. Then, we compared the ANNA dataset with the COCO dataset and presented the evaluation of the IOU metrics for multiple object detection on the roadways. This approach aims to provide autonomous cars with enough knowledge to travel safely in excessive traffic conditions in Bangladesh.

\section{Related Work}
Self-driving technology~\cite{p6} is a long-term endeavor to improve safety by automating vehicle control. The autonomous car is a decision-making system that gathers data from numerous sources, such as cameras, LiDAR, RADAR, and other sensors~\cite{p17}. With the increasing use of Deep Neural Networks (DNN) models for object detection, it can be stated that DNN is capable of processing information in real-time and can transmit information to cloud storage as well as to automated vehicles~\cite{p4}. Deep Learning involves training and testing on labeled data, which can be labeled in the case of self-driving cars and annotated using ground-truth bounding boxes~\cite{p4}. To recognize and classify objects, machine learning (ML) methods such as Naive Bayes, Support Vector Machines, and CNN-based DL models are utilized~\cite{p1}. CNN-based systems can play a significant role in pedestrian, lane, and redundant object detection at moderate distances~\cite{p2, p17}. CNN models have achieved 100\% classification rates on datasets such as ImageNet\cite{p20}. In YOLO, multi-class object detection can be performed in real-time~\cite{p2}. YOLO is a one-stage detector with a high detection speed that directly predicts the bounding box parameters for the objects in the image and classifies these objects simultaneously~\cite{p9, p18}.

The camera can be a cost-effective solution compared to LIDAR for basic detection needs in autonomous driving~\cite{p42}. Monocular cameras provide detailed information in the form of pixel intensities, which, at a larger scale, reveal shape and texture properties~\cite{p3}. Using a monocular camera, it is possible to estimate the distance of objects in front of autonomous vehicles~\cite{p42, distance1}. Monocular image-based 3D object detection methods utilize RGB images to predict 3D object bounding boxes~\cite{p3}.

The Cityscapes dataset contains pixel-level semantic labeling for 25k images depicting urban scenarios from 50 cities~\cite{p31}. The ApolloScape dataset~\cite{p28} provides 140k labeled images of street views for lane detection, car detection, semantic segmentation, and more. It enables performance evaluation across different times of day and weather conditions. The WoodScape dataset for AV~\cite{p32} offers 10k images captured from four dedicated fisheye cameras with semantic annotations for 40 classes. The nuScenes dataset~\cite{p21} addresses the need for multimodal data and includes 23 categories such as vehicles, pedestrians, mobility devices, and other objects. It comprises 1000 scenes, each 20 seconds long, with 3D bounding box annotations for 23 classes and 8 attributes, captured using six cameras, five radars, and one lidar with a full 360-degree field of view. The Oxford RobotCar dataset~\cite{p29} route in Oxford, UK and consists of 32 traversals in different traffic, weather, and lighting conditions totalling 280 km of urban driving resulting 4.7TB of the . The Lyft dataset~\cite{p24} evaluates the prediction efficiency of DL models by calculating the Root Mean Square Error and provides high-definition scene data with bounding boxes and class probabilities. It includes over 1000 hours of training data along a single route. The MS COCO~\cite{p27} dataset consists of 328k images with 80 labels representing commonly available objects. The Coyote dataset~\cite{p25} comprises 894 real-world photographs with over 1700 ground-truth labels, grouped into six broad categories. The WildDash~\cite{p26} dataset contains 1800 frames that address natural risks in images, such as distortion, overexposure, windscreen, road conditions, weather, and lighting variations, from diverse geographic locations. 

The KITTI dataset, which consists of 13k images of raw data, was obtained by annotating 400 dynamic scenes using detailed 3D CAD models for all moving vehicles~\cite{p22}. The Virtual KITTI dataset~\cite{p33} reconstructs the KITTI environment using a gaming engine and 3D models to create labeled data with different variations. CARLA~\cite{p30} is an open-source autonomous driving simulation tool that allows for customizable environmental setups and sensor configurations. Experimental results show that simulator-generated datasets yield similar DNN prediction errors to real-world datasets~\cite{p23}. However, it is important to note that simulator-generated datasets may not fully represent real-life driving environments, and safety violations were identified~\cite{p23}. Therefore, in this paper, a custom-made dataset collected from a real-life environment was utilized.

Since our dataset includes a diverse range of objects beyond just vehicles, we leverage the MS COCO dataset~\cite{p27} as the foundation for our trained classifiers and for comparative evaluation. While some datasets offer precise pixel-level semantic segmentations, they may not encompass the wide range of scenarios present in our ANNA dataset. The use of the ANNA dataset is intended as an initial step to evaluate computer vision for autonomous vehicles in challenging traffic conditions and with uncommon vehicles. Furthermore, this dataset can be expanded to train computer vision systems that are more robust and capable of handling adversarial situations, ultimately enhancing car safety.

\section{Overview of the ANNA dataset}  %section
  
\begin{figure}
    \centering
    \includegraphics[width=9cm]{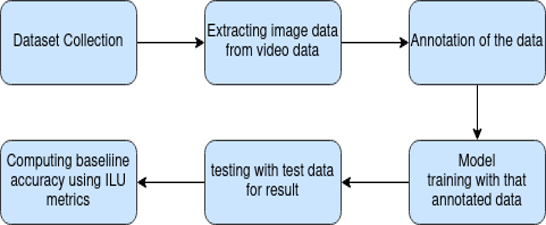}
    \caption{Overview of the ANNA dataset}
    \label{fig:methodology1}
\end{figure}

A six-step process was used to construct a multiple object detection model for autonomous vehicles from the perspective of Bangladesh (Fig. \ref{fig:methodology1}). To develop autonomous vehicles equipped with mobile cameras, the steps are described in the following subsections:

\subsection{Data Collection} %subsection

\subsubsection{Planning the route}
Specific routes were planned for data collection, ensuring coverage of various types of areas including urban and rural roads, freeways, and crossroads. Our data was collected from multiple roads, including those in Mirpur (Dhaka), Bashundhara Residential Area (Dhaka), Jhenidah (Khulna), and other locations.

\subsubsection{Setting up the vehicle}
The vehicle was equipped with a mobile camera capable of recording high-definition video, which was mounted in a position to provide a clear view of the surroundings. Our dataset was collected using various devices including the iPhone 13 Pro, Huawei Nova 3i, and Samsung Galaxy A12 (48 Megapixels).

\subsubsection{Scene selection}
The vehicle was operated in a manner that allowed for the recording of a wide range of scenes, including different traffic situations, weather conditions, and lighting conditions. Over 20 videos were collected from low-traffic areas, and 1800 images were manually selected for annotation. When choosing scenes for autonomous cars equipped with mobile cameras, it is important to consider various scenarios, including high-traffic areas such as intersections and construction sites, rare vehicle classes, and potentially dangerous situations. To ensure that the dataset included a diverse range of vehicles commonly seen in Bangladeshi neighborhoods, footage of several vehicle types, including cars, buses, CNGs (compressed natural gas-driven three-wheelers), vans (rickshaw vans), easy bikes, motorbikes, rickshaws, and others, was collected.

\begin{figure}
    \centering
    \includegraphics[width=9cm]{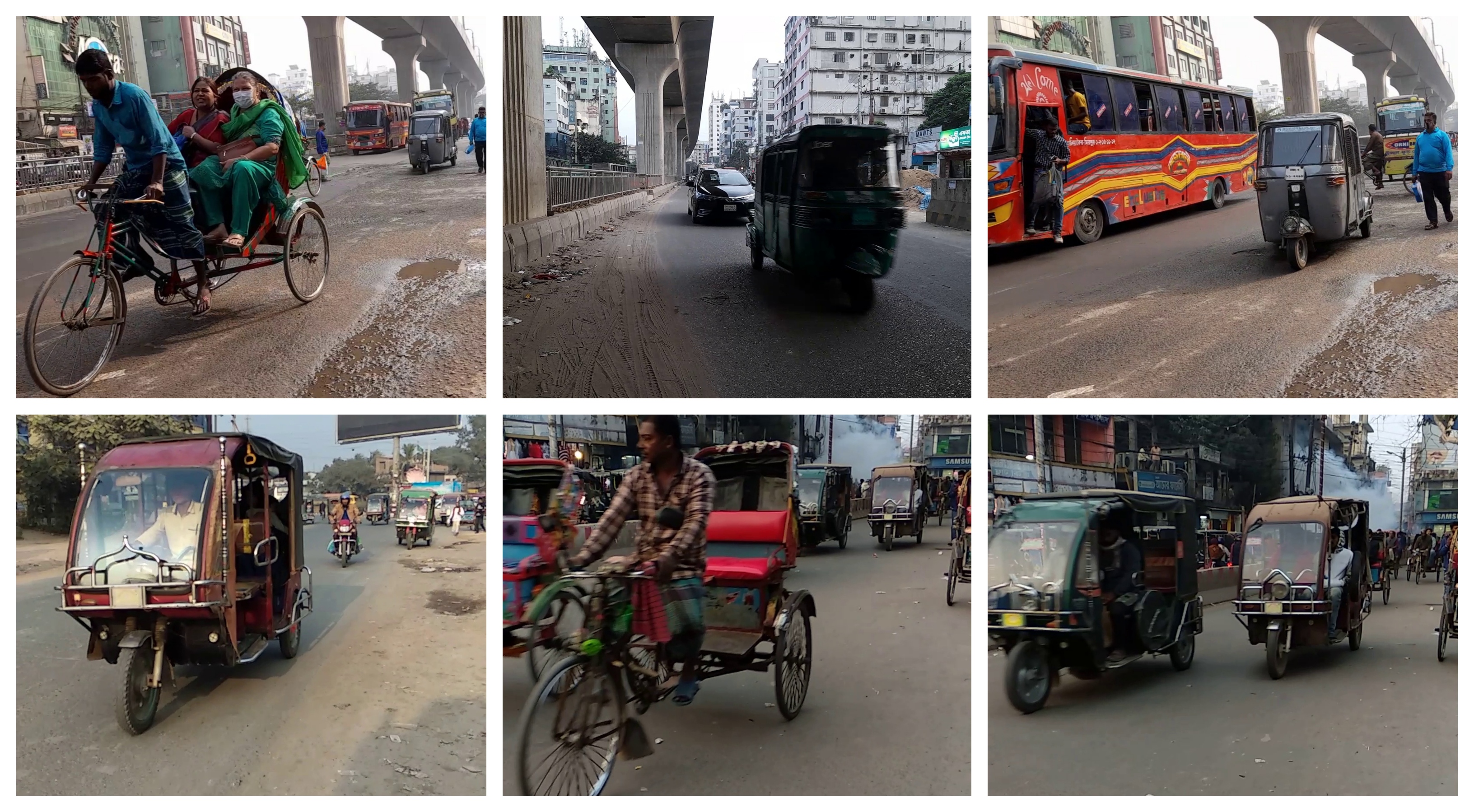}
    \caption{Extracted image dataset}
    \label{fig:sys-collage2}
\end{figure}

\subsection{Data Extraction} % Subsection
Image data extraction typically involves reading image files and converting them into a format that can be utilized by machine learning models or other computer vision algorithms (Fig. \ref{fig:sys-collage2}. The following provides a high-level overview of the process:

\subsubsection{Read the video file}
The first step is to use a video I/O library or API to read the video file. The video frames are accessible through the library as a series of images.

\subsubsection{Extract individual frames}
Once the video is loaded, individual frames are extracted from the video sequence. This is done by looping over the video frames using a customized Python script.

\subsubsection{Convert the frames to a standard format}
After extracting the frames, they are converted into a standard picture format that can be used by machine learning models or other algorithms.

\subsection{Data Annotation}  %subsection
\begin{figure}
    \centering
    \includegraphics[width=9cm]{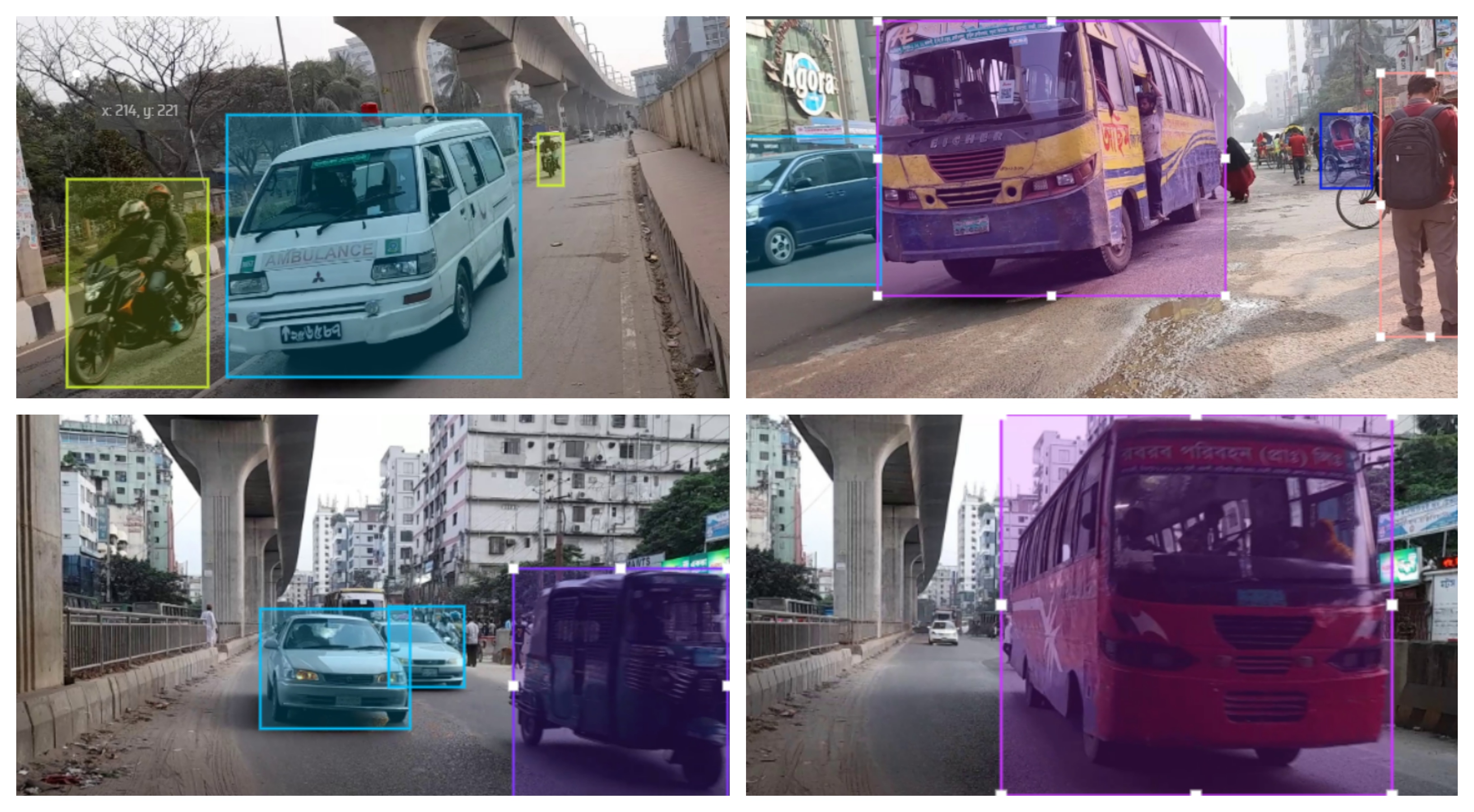}
    \caption{Data annotation}
    \label{fig:sys-collage3}
\end{figure}

To provide detailed information about the objects identified in each frame, the video data collection needs to be annotated. Computer vision algorithms are trained using this annotated data to detect objects in video frames. For this project, a specific set of 10 scenes and 1800 selected images were manually chosen for annotation (Fig \ref{fig:sys-collage3}). The annotation process involved labeling eight classes, including humans, cars, buses, rickshaws, bikes, CNGs, bicycles, easy bikes, and vans (rickshaw vans). Expert annotators were employed to annotate objects continuously throughout each scene. An online AI software called \href{https://www.makesense.ai}{makesense.ai} was utilized for the annotation process.

By involving experts in the annotation process and implementing multiple validation steps, highly accurate annotations were achieved. Labeling objects in the captured video data is a necessary step in the data annotation process for an autonomous vehicle implementing the YOLOv5 computer vision algorithm. The system is trained to identify objects in each video frame, such as other vehicles, pedestrians, and more, using the annotated data. The location and class of each object in every frame are manually marked as part of the data annotation process. The dataset is available at \href{The dataset is available at https://github.com/MahediKamal/ANNA-A-Deep-Learning-Based-Dataset-in-Heterogenous-Traffic-for-Autonomous-Vehicles}{https://github.com/MahediKamal/ANNA-A-Deep-Learning-Based-Dataset-in-Heterogenous-Traffic-for-Autonomous-Vehicles}.

\subsection{Model Training with Data}  %subsection
\subsubsection{Algorithms and frameworks used}
After the data has been annotated, it can be utilized to train the YOLOv5 object detection model. The model is then tested on fresh, unlabeled data to evaluate its accuracy and make any necessary adjustments. YOLOv5, a well-known deep learning (DL) library, can be employed for training object detection models. Torch, a DL framework, provides a wide range of tools, libraries, and support for constructing, refining, and deploying DL models, including computer vision tasks like object classification. When used in conjunction with YOLOv5, the Torch framework enables network improvement and reduction of error between the predicted and actual object positions. Once the model is trained, it can be applied in autonomous driving applications.

\subsubsection{Training the model}    %subsection
While constructing the model, the data was divided into two sets: (1) test data and (2) training data. Each set consisted of images along with their corresponding annotations. The model was trained for 100 epochs with a batch size of 4. In YOLOv5, weight files and pointer files are utilized to store the learned parameters of the neural network model during the training process. Pointer files (.yaml) are text files that specify the structure of the neural network model and the location of the weight files on disk. These pointer files contain metadata about the network architecture, such as the number and size of layers, as well as the location of the weight files that contain the learned parameters.

Weight files (.pt) contain the actual numerical values of the trained parameters of the neural network model. These weight files are binary files that can be loaded into the model during inference to apply the learned parameters to new input data. During training, the weights are adjusted to minimize the error between the predicted and actual object locations and are then stored in the weight files. Together, the weight files and pointer files enable the YOLOv5 algorithm to accurately detect objects in real-world scenarios.

\subsection{Model Deployment}       %subsection

After the model has been trained and tested, it can be deployed for use in an autonomous car or another system to detect objects in a live stream. During deployment, the YOLOv5 model takes a video frame as input and applies a series of CNN layers to generate a set of bounding boxes around objects in the frame. Each bounding box is assigned a class label and a confidence score, indicating the model's confidence in the presence of an object of that class within the bounding box. The model then applies non-maximum suppression to eliminate overlapping bounding boxes and outputs the final set of object detections.

\subsection{Model Testing and Evaluation}       %subsection

\begin{figure}
    \centering
    \includegraphics[width=9cm]{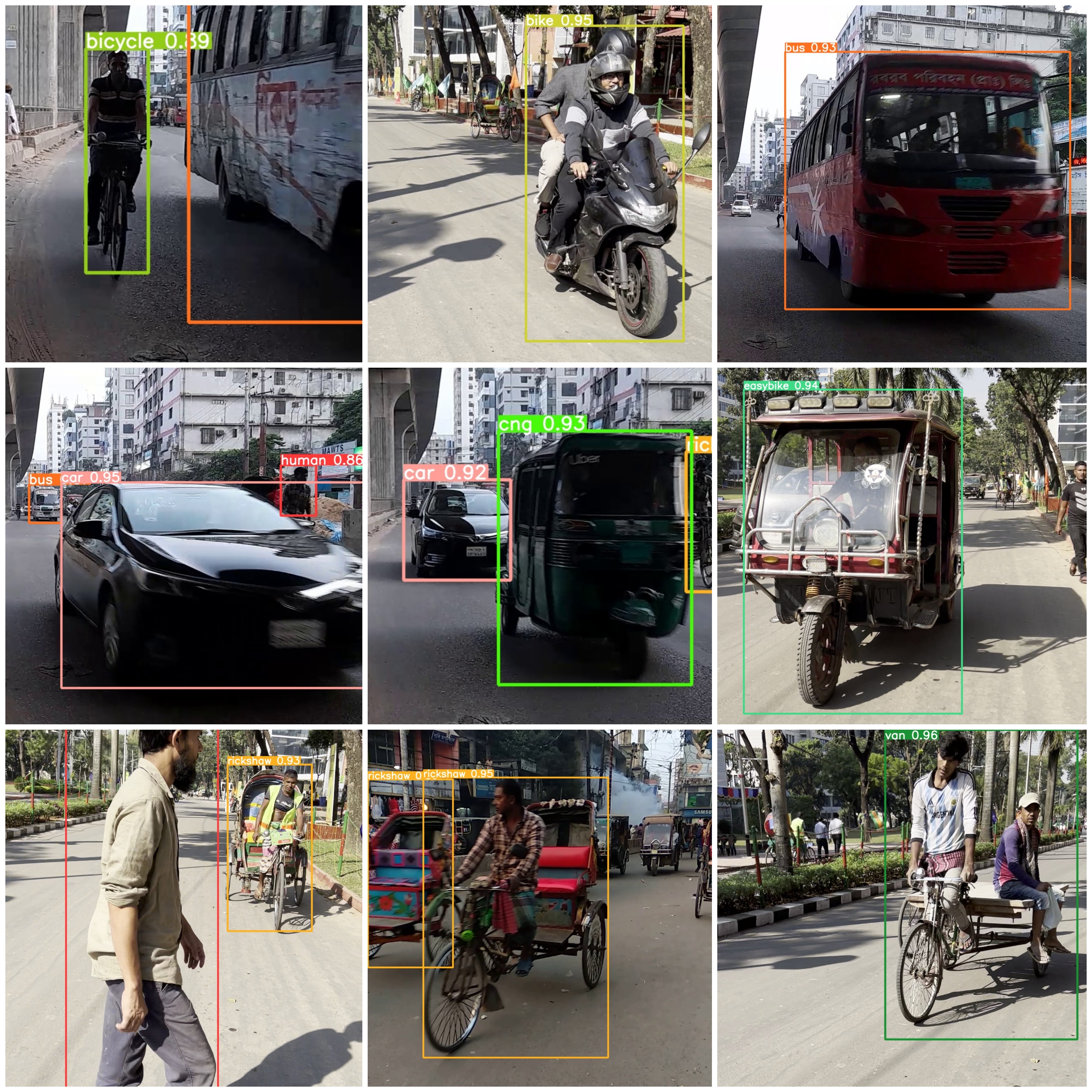}
    \caption{Detection of common and rare vehicles}
    \label{fig:sys-collage4}
\end{figure}

The Mean Average Precision (MAP) metric was utilized to evaluate the performance of our multiple object detection model (Fig. \ref{fig:sys-collage4}). A higher MAP value indicates better performance. MAP calculates the average precision across all object categories. We evaluated the model's performance on the testing set, which consisted of images that were not used during training. For each image in the testing set, we computed the average precision for each object category and then averaged the results across all object categories to obtain the MAP score. To calculate the average precision for each object category, we initially computed accuracy and recall values for various confidence thresholds. True positive, false positive, and false negative values were determined by comparing the projected bounding boxes with the ground truth bounding boxes. The precision and recall data were used to calculate the average precision for each category.

\section{Result Analysis}    %section

A comparison was conducted between the custom ANNA dataset and one of the most popular datasets, MS-COCO, for multiple object detection evaluation. Initially, a model trained on the MS-COCO dataset was applied to the dataset collected from Bangladeshi roads to generate predictions. To ensure that only objects from the selected 9 classes were detected by the model, some modifications were made during the prediction process. Subsequently, another model was trained on the ANNA dataset, and using this model, predictions were made on the dataset collected from Bangladesh. Finally, a comparison was made between the two sets of predictions and the ground truth. For this, the concept of IOU (Intersection Over Union) was used. IOU is a measure based on the Jaccard Index that evaluates the overlap between two bounding boxes. It requires a ground truth bounding box and a predicted bounding box. By applying the IOU we can tell if a detection is valid (True Positive) or not (False Positive). Based on IOU performance metrics; average precision (AP) is calculated. Average precision (AP) is a commonly used evaluation metric in machine learning for object detection or image segmentation tasks. It measures the average precision of a machine learning model across all possible values of the threshold used for making predictions. The AP value provides an overall assessment of the model's performance. The comparison table below presents the results obtained from the MS COCO and ANNA datasets.

\begin{table}
    \centering
    \begin{tabular}{ | p{2.3cm} | p{2.3cm} | p{2.3cm} | }
         \hline
         \textbf{Object Class (Vehicle names)} & \textbf{Average Precision (AP) for ANNA dataset (\%)} & \textbf{Average Precision (AP) for COCO dataset (\%)}  \\ 
         \hline
         Human (0) & 90.73 & 92.26 \\ 
         \hline
         Car (1) & 96.05 & 95.05  \\ 
         \hline
         Bus (2) & 87.7 & 82.25  \\
         \hline
         Rickshaw (3) & 96.92 & 0.00  \\
         \hline
         Bike (4) & 95.59 & 89.50  \\
         \hline
         CNG (5) & 95.40 & 0.00  \\
         \hline
         Bicycle (6) & 93.20 & 82.34  \\
         \hline
         Easy Bike (7) & 97.95 & 0.00  \\
         \hline
         Van (8) & 91.11 & 0.00  \\
         
         \hline
    \end{tabular}
    \caption{Comparison Between ANNA and COCO dataset}
    \label{tab:Table 1}
\end{table}

Considering the 9 classes of vehicles specified earlier; the mean average precision (MAP) for the ANNA dataset is 93.85\%, while for the COCO dataset, it is 49.04 \%. The MAP value significantly decreases for the COCO dataset due to its inability to detect rickshaws, CNGs, easy bikes, and vans (rickshaw vans). However, since the ANNA dataset includes data specifically for these vehicles, a model trained on ANNA can accurately detect them.

\section{Conclusion}        %section

In this paper, the ANNA dataset is presented which was collected from busy and public roads of Bangladesh and also, tracking tasks, metrics, baselines, and results are shown. This is the first dataset where uncommon vehicles of Bangladesh which can not be found in any of the AV datasets, are collected to test on public roads. Our dataset is only based on a mobile camera which completes the Waymo challenge containing “Real-time 2D Detection”~\cite{p43, waymo2021} and "3D Camera-Only Detection"~\cite{waymo2022}. The IOU metrics are introduced that balance all components of detection performance in order to encourage research on 3D object detection for AVs in Bangladesh. Further work will be done on increasing the number of images by increasing the challenging scenarios; e.g. occluded objects such as a person occluded under an umbrella, the field of views from each object, humans or other animals in vehicles, vehicles in bad weather conditions and by capturing it from a camera within a vehicle itself.

\bibliographystyle{IEEEtran}
\bibliography{main}

\end{document}